\pdfoutput=1
\documentclass{article}

\def\arxivversion{1}

\usepackage{microtype}
\usepackage{graphicx}
\usepackage{booktabs}
\usepackage{amsmath}
\usepackage{amssymb}
\usepackage{xspace}
\usepackage{enumitem}
\usepackage[table]{xcolor}
\usepackage{algorithm}
\usepackage{algorithmic}
\usepackage{flushend}
\usepackage{xr-hyper}
\usepackage{hyperref}

\expandafter\def\expandafter\UrlBreaks\expandafter{\UrlBreaks
  \do\a\do\b\do\c\do\d\do\e\do\f\do\g\do\h\do\i\do\j\do\k\do\l\do\m\do\n\do\o\do\p\do\q
  \do\r\do\s\do\t\do\u\do\v\do\w\do\x\do\y\do\z\do\A\do\B\do\C\do\D\do\E\do\F\do\G\do\H
  \do\I\do\J\do\K\do\L\do\M\do\N\do\O\do\P\do\Q\do\R\do\S\do\T\do\U\do\V\do\W\do\X\do\Y
  \do\Z\do0\do1\do2\do3\do4\do5\do6\do7\do8\do9}

\ifdefined\arxivversion
\usepackage[accepted]{mlsys2025}
\makeatletter
\renewcommand{\Notice@String}{Preprint.}
\makeatother
\else
\usepackage{mlsys2025}
\fi

\usepackage[capitalise,noabbrev]{cleveref}

\usepackage[caption=false,font=footnotesize,captionskip=1pt,farskip=2pt,nearskip=0pt]{subfig}

\usepackage{tikz}
\usetikzlibrary{positioning,arrows.meta,fit,backgrounds,calc,shapes.geometric,patterns,
  decorations.pathreplacing}

\newcommand{\sys}{FactorSpec\xspace}
\newcommand{\xg}{XGrammar\xspace}

\newcommand{\code}[1]{\texttt{#1}}
\newcommand{\takeaway}[1]{%
  \par\addvspace{3pt}\noindent
  {\setlength{\fboxsep}{2pt}\setlength{\fboxrule}{0.3pt}%
   \fbox{\parbox{\dimexpr\linewidth-2\fboxsep-2\fboxrule\relax}{%
     \textbf{Takeaway.} #1}}}%
  \par\addvspace{2pt}}

\newif\ifprovisional
\provisionalfalse

\definecolor{cXg}{HTML}{8C8C8C}       
\definecolor{cCombined}{HTML}{006D2C} 
\definecolor{cService}{HTML}{FFFFFF}  
\definecolor{cEngine}{HTML}{F3F3F3}   
\definecolor{cLine}{HTML}{3A3A3A}

\tikzset{
  >={Stealth[length=3.5pt]},
  tok/.style={draw=cLine!70, rounded corners=1pt, inner sep=1.5pt, minimum height=0.34cm,
    font=\tiny\ttfamily, line width=0.4pt},
  det/.style={tok, fill=cCombined!18, draw=cCombined},
  cand/.style={tok, fill=white, draw=cCombined, line width=0.6pt},
  graft/.style={tok, fill=cLine!8, draw=cLine!60, dashed},
  edge/.style={line width=0.5pt, cLine!80},
  part/.style={draw=cLine!60, fill=white, rounded corners=1.5pt, line width=0.4pt,
    align=center, font=\tiny, inner sep=2pt},
  flow/.style={->, line width=0.6pt, cLine},
  rowflow/.style={->, line width=1pt, cCombined},
  num/.style={circle, fill=cLine, text=white, font=\tiny\bfseries, inner sep=0.5pt},
  lane/.style={font=\tiny, anchor=west, text=cLine},
  gpu/.style={fill=cXg!45, draw=cXg!80!black, line width=0.4pt},
  cpu/.style={fill=cEngine, draw=cLine!60, line width=0.4pt},
  work/.style={fill=cCombined!18, draw=cCombined, line width=0.4pt},
  idle/.style={pattern=north east lines, pattern color=cLine!60, draw=cLine!70,
    line width=0.4pt},
  barlabel/.style={font=\tiny, inner sep=0.5pt},
}

\newcommand{\circled}[1]{\tikz[baseline=(n.base)]\node[num, font=\scriptsize\bfseries,
  inner sep=0.6pt] (n) {#1};}

\setlist{leftmargin=*, itemsep=1pt, topsep=2pt}

\makeatletter
\long\def\@makecaption#1#2{%
  \vskip 5pt
  \baselineskip 11pt
  \setbox\@tempboxa\hbox{#1. #2}%
  \ifdim \wd\@tempboxa >\hsize
    \sbox{\newcaptionbox}{\small\sl #1.~}%
    \newcaptionboxwid=\wd\newcaptionbox
    \usebox\newcaptionbox {\footnotesize #2}%
  \else
    \centerline{{\small\sl #1.} {\small #2}}%
  \fi
  \ifx\@captype\@table\par\vskip 4pt\fi}
\def\@table{table}
\makeatother


\mlsystitlerunning{FactorSpec: Serving Exact Relational Constraints as Speculative Drafts}

\begin{document}

\twocolumn[
\mlsystitle{FactorSpec: Serving Exact Relational Constraints as Speculative Drafts}

\begin{mlsysauthorlist}
\mlsysauthor{Jianchang Su}{uconn}
\mlsysauthor{Dawei Xiang}{uconn}
\mlsysauthor{Wei Zhang}{uconn}
\end{mlsysauthorlist}
\mlsysaffiliation{uconn}{University of Connecticut, Storrs, CT, USA}
\mlsyscorrespondingauthor{Wei Zhang}{wei.13.zhang@uconn.edu}

\mlsyskeywords{LLM serving, constrained decoding, speculative decoding, structured output}

\vskip 0.3in

\begin{abstract}
Speculative decoding speeds up LLM serving by drafting tokens that the target model
verifies in one pass, and its gain grows with the drafts the target accepts. Structured
outputs should be an ideal case, because much of a record follows from relations among its
fields: an itinerary's days follow from its cities' stays and flights. Serving engines,
however, draft such text by proposing it from the prompt or a learned head, and their grammar
masks enforce only syntax, so the model must satisfy the relations, which it violates: on
NATURAL PLAN's trip problems, Qwen3-8B writes a valid itinerary for 10\%, and retries that
name the violated relation raise this to 14\%. We present
\sys, which serves exact relational constraints as speculative drafts. \sys compiles a JSON
Schema and its relations into a factor graph with an exact feasibility check, drafts every
token the constraint determines as certain, verifies each request at a width chosen for it,
and computes the masks in worker processes outside the scheduler while the GPU runs. In
SGLang, its drafts yield up to 10.5 tokens per verify step, up to 1.9$\times$ those of the
fastest n-gram drafter, and it decodes up to 1.43$\times$ faster than speculators tuned per
load, at a grammar mask's cost; it also leads in vLLM. Because every record is also valid, it serves up
to 13.9$\times$ the valid records per GPU-second of syntax-only baselines.
\end{abstract}

]

\printAffiliationsAndNotice{}

\section{Introduction}
\label{sec:intro}

Speculative decoding~\cite{leviathan2023fast,chen2023accelerating} speeds up LLM serving by
drafting tokens that the target verifies in one forward pass, and its gain grows with the
drafts the target accepts. Structured outputs, such as tool calls and plans, should be an
ideal case, because much of a record follows from relations among its fields: a refund's
item must belong to its order, and an itinerary's days follow from its cities' stays and
the direct flights between them. Serving engines such as SGLang~\cite{zheng2024sglang} and
vLLM~\cite{kwon2023efficient}, however, check only syntax, and the model violates the
relations: on 999 trip problems of three to seven
cities from NATURAL PLAN~\cite{zheng2024natural}, Qwen3-8B writes a valid itinerary for 10\%, and retries that
name the violated relation reach 14\% (\cref{sec:observations}).

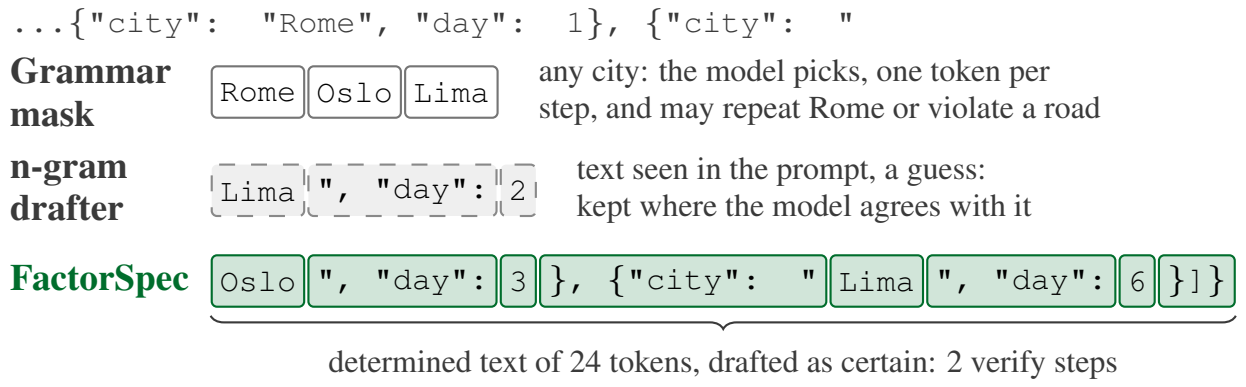
\begin{figure}[t]
\centering
\begin{tikzpicture}[node distance=0.4pt]
\node[anchor=west, font=\tiny\ttfamily, text=cLine] at (0,0)
  {\ldots\{"city": "Rome", "day": 1\}, \{"city": "};
\node[anchor=west, font=\scriptsize\bfseries, text=cLine, align=left] at (0,-0.45)
  {Grammar\\mask};
\node[tok, anchor=west] (a1) at (1.45,-0.45) {Rome};
\node[tok, right=of a1] (a2) {Oslo};
\node[tok, right=of a2] (a3) {Lima};
\node[anchor=west, font=\tiny, text=cLine, align=left, right=4pt of a3]
  {any city: the model picks, one token per\\step, and may repeat Rome or violate a road};
\node[anchor=west, font=\scriptsize\bfseries, text=cLine, align=left] at (0,-1.1)
  {n-gram\\drafter};
\node[graft, anchor=west] (n1) at (1.45,-1.1) {Lima};
\node[graft, right=of n1] (n2) {", "day":};
\node[graft, right=of n2] (n3) {2};
\node[anchor=west, font=\tiny, text=cLine, align=left, right=4pt of n3]
  {text seen in the prompt, a guess:\\kept where the model agrees with it};
\node[anchor=west, font=\scriptsize\bfseries, text=cCombined] at (0,-1.7) {\sys};
\node[det, anchor=west] (b1) at (1.45,-1.7) {Oslo};
\node[det, right=of b1] (b2) {", "day":};
\node[det, right=of b2] (b3) {3};
\node[det, right=of b3] (b4) {\}, \{"city": "};
\node[det, right=of b4] (b5) {Lima};
\node[det, right=of b5] (b6) {", "day":};
\node[det, right=of b6] (b7) {6};
\node[det, right=of b7] (b8) {\}]\}};
\draw[decorate, decoration={brace, mirror, amplitude=2.5pt}, cLine, line width=0.4pt]
  ([yshift=-1.5pt]b1.south west) -- ([yshift=-1.5pt]b8.south east)
  node[midway, below=3pt, font=\tiny, text=cLine]
  {determined text of 24 tokens, drafted as certain: 2 verify steps};
\end{tikzpicture}
\caption{Decoding the tour of \cref{fig:declare} after its first stop, three ways. A grammar
mask admits every city, and the model chooses token by token. An n-gram drafter proposes
text it has seen, which the target keeps where it agrees. \sys's relations admit only Oslo on day 3, then Lima on day 6; it drafts the determined
tokens, the target accepts them all, and the record completes in two steps.}
\label{fig:mechanism}
\end{figure}

Existing approaches each cover part of this. Grammar
engines~\cite{dong2025xgrammar,willard2023efficient} answer in microseconds but enforce
syntax, and an enum schema closes the gap where a record copies values from its prompt.
Semantic constrained decoders~\cite{scholak2021picard,poesia2022synchromesh,mundler2025type}
check relations locally, which on joint relations leads a decoder to infeasible prefixes:
served in an engine, arc consistency leaves 12 to 64\% of planning records incomplete.
Solver
pipelines~\cite{liu2023llmp,pan2023logiclm} compute a valid answer outside the model, which
then still decodes it token by token. The engines' drafters~\cite{li2025eagle} propose
likely or repeated text, and jump-forward decoding~\cite{zheng2024sglang} skips the tokens
a grammar determines; the text the relations compute, such as the day a stay must end, lies
outside all of them.

Drafting from the relations themselves would serve both speed and validity, but it poses
two systems problems. The first is speed at serving load: a drafter's gain shrinks as the
batch grows~\cite{liu2026speculative}, as one width per step spends slots on guesses. The
second is cost: an exact check over joint relations, such as an all-different over cities
or arithmetic over days, is far heavier than a grammar mask, and on the scheduler thread
it would delay every request of the step.

Our insight is that an exact constraint is itself a drafter (\cref{fig:mechanism}). It knows
which positions it determines, so it can propose those tokens, and it knows for each request how many of its
drafts the target will accept. It also knows its next state before the target samples a
determined token, so it can compute ahead while the device runs. The constraint thus turns its cost into a source of speed.

\sys builds this insight into speculative decoding for structured outputs. A client declares
a record as a JSON Schema and a list of relations, which compile to a factor graph with an
exact feasibility check and a token-level structure engine that admits a token only if it
extends to a valid record. Against the speed problem, speculative drafters propose the
constraint's determined runs as certain drafts, alone or grafted with the engine's n-gram
corpus or EAGLE-3 head, and verify each request at a width chosen for it. Against the cost
problem, a constraint service of worker processes holds each request's session outside the
scheduler process and writes mask rows into shared memory while the forward pass runs.

We evaluate \sys in SGLang and vLLM on six tool-calling and planning workloads and find that
exactness is both fast and cheap. On Qwen3-8B, each verify step accepts up to 10.5 tokens,
up to 1.9$\times$ those of the fastest n-gram shape, because a determined run is accepted
whole. That acceptance length lets the combined drafter match or outrun the engine's
speculator tuned per load at 8, 64 and 128 requests in flight (\cref{fig:spec}), by up to
1.43$\times$ where the relations compute the record and 1.01 to 1.12$\times$ where the
model chooses most of it. The masks alone run at 0.95 to 1.07 of \xg's throughput up to 64
in flight, because the workers overlap the GPU, so the cost of exactness is host CPU.
Because every record is also valid, the gain in usable output is far larger: \sys serves up
to 13.9$\times$ the valid planning records per GPU-second of syntax-only baselines. A
solver makes every record valid too, so against it the margin is speed, 1.04 to
1.46$\times$; where gold follows a fixed objective the solver's records match gold more
often. These results hold, as leads or ties, across two engines, four models, tensor parallelism
and sampling.

This paper makes the following contributions.
\begin{itemize}
\item The constraint as a lossless drafter that knows which drafts are certain, with
  per-request verify widths, where engines verify a step at one width
  (\cref{sec:drafts,sec:proposer,sec:width}).
\item Exact per-token enforcement of joint relations, where semantic decoders check local
  ones: a memoized feasibility check over factor graphs, a structure engine that keeps
  every admitted token completable, and a JSON Schema frontend (\cref{sec:engine,sec:tokens}).
\item A constraint service outside the scheduler process, where grammar engines run inside
  it: worker processes write shared-memory mask rows and compute ahead
  (\cref{sec:service}).
\item An evaluation in two engines, on three model sizes, two families and six workloads,
  against speculators tuned per load, enum schemas, retries, reasoning, solvers and
  enumerated schemas (\cref{sec:eval}).
\end{itemize}

\section{Background and Motivation}
\label{sec:background}

\subsection{Speculative decoding in a serving engine}

A serving engine runs a scheduler loop that admits requests, forms a batch, launches the
target's forward pass and processes the previous iteration's results. Under speculative
decoding, a drafter proposes several tokens per request, and the target verifies them in
one forward pass~\cite{leviathan2023fast,chen2023accelerating}, a tree per request in
SGLang~\cite{miao2024specinfer} and a chain in vLLM. Drafts come from a learned head such as
EAGLE-3~\cite{li2025eagle} or from an n-gram corpus of earlier text. The tokens a verify step yields per request, the
\emph{acceptance length}, set the speedup when the batch is small. As the batch grows, each
verify slot also costs time, and the engine verifies every request of a step at the same
width, so the width suits the batch on average.

\subsection{Structured generation}

For a constrained request the engine keeps a \emph{grammar object} with three calls: \code{fill}, which writes the
request's row of a vocabulary bitmask before sampling; \code{accept}, which advances the
object by a sampled token; and \code{rollback}. A grammar engine such as
\xg~\cite{dong2025xgrammar} compiles a schema into an automaton and answers these calls in
C++ in a few microseconds. The engine makes the calls on its scheduler thread, one request
after another, so their cost adds to every step: negligible at microseconds per call, and
significant for a heavier check. Under speculation, the grammar supplies a mask row per
draft node.

\subsection{Relations as a factor graph}
\label{sec:factorgraph}

A record's fields that take finitely many values are variables, and a relation among
fields is a factor, a Boolean table over the variables in its scope. The \emph{support} of a
variable under evidence is the set of its values that extend to a full assignment
satisfying every factor. A mask needs exactly this set: a value outside it reaches an
infeasible prefix, and every value inside it can still complete the record. Bucket elimination computes
the support
exactly~\cite{dechter1999bucket} at a cost exponential in the induced width of its
elimination order. Such relations are costly to state as a grammar: to admit references to
just the $k$ of $d$ identifiers a record declared earlier, an automaton must
remember which $k$ were declared, up to $\binom{d}{k}$ states, while a factor graph keeps
the identifiers as variables and relates each reference directly.

\subsection{Motivation}
\label{sec:observations}

\begin{table}[t]
\centering
\caption{The share of output tokens of 200 gold records per workload that the spec
determines, admitting one token given the prefix, by source: \emph{syntax} the grammar
alone fixes; \emph{copied} text appears in the prompt; \emph{computed} is the rest, an
upper bound, since a run with any computed value counts whole. Trip: seven-city problems.}
\label{tab:census}
\footnotesize
\setlength{\tabcolsep}{4pt}
\begin{tabular}{@{}lrrrr@{}}
\toprule
Workload & Determined & Syntax & Copied & Computed \\
\midrule
Ledger & 0.768 & 0.262 & 0.445 & 0.061 \\
$\tau$-bench retail & 0.858 & 0.154 & 0.704 & 0.000 \\
BFCL multi-turn & 0.702 & 0.294 & 0.409 & 0.000 \\
Trip planning & 0.9998 & 0.000 & 0.000 & 0.9998 \\
Meeting planning & 0.895 & 0.039 & 0.063 & 0.793 \\
Calendar planning & 0.961 & 0.071 & 0.838 & 0.053 \\
\bottomrule
\end{tabular}
\end{table}

\begin{figure}[t]
\centering
\includegraphics{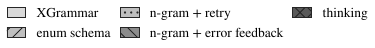}\\
\includegraphics{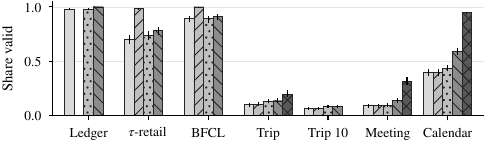}
\caption{When the model must satisfy the relations, trip and meeting records violate them,
and retries and reasoning repair few. The share of each full set at 64 in flight whose record is
valid, Qwen3-8B, 95\% Wilson intervals; Trip 10 is all 1,599 trip problems; enum schemas on
every workload except the ledger; thinking on 64 problems of each planning workload at 8 in
flight, range of three runs.}
\label{fig:motivation}
\end{figure}

We measured Qwen3-8B on SGLang with workloads from tool calling and planning
(\cref{sec:eval-setup}). Three observations motivate the design: an opportunity (O1), a
requirement (O2) and a challenge (O3), each met by one part of \cref{sec:design}.

\textbf{O1: The relations determine most of the record, and the engine's speculators
propose little of it at load.} \Cref{tab:census} splits the determined tokens of gold
records by source. On the tool calls the determined text is mostly copied from the prompt,
which a corpus or a prompt matcher can also propose. On trip planning all of the itinerary
and on meeting planning 79\% of the schedule is computed from the problem, so those tokens
lie outside the reach of both drafters. The engine's speculators miss these tokens, because their gain falls with the batch: tuned
per load, they run at 1.38 to 2.86$\times$ \xg at 8 requests in flight but only 0.94 to
1.17$\times$ at 128 (\cref{fig:spec}), because a step verifies every request at one width.
The constraint should draft what it determines and size each request's verify by it
(\cref{sec:drafts,sec:proposer,sec:width}).

\textbf{O2: When the model must satisfy joint relations, it violates them, and retries and
reasoning repair few.} The drafts must also be valid. On tool calls, whose relations are
mostly memberships, \xg alone on the ledger and an enum schema on retail and BFCL make 0.98
to 1.00 of the records valid (\cref{fig:motivation}), because a membership is a per-field
check. On trip and meeting planning, whose relations join fields, an enum schema leaves
validity where \xg's is (\cref{fig:motivation,fig:breakdown}(b)), and the best baseline makes under a third of the records valid.
Retries succeed on a minority of problems, whether the model samples three attempts,
receives feedback or reasons for thousands of tokens. A weaker check misses the cycles too:
arc consistency leads 12 to 64\% of planning records to infeasible prefixes
(\cref{sec:eval-enforce}), because local propagation sees each constraint in isolation. The
relations must be enforced exactly while the model decodes (\cref{sec:engine,sec:tokens}).

\textbf{O3: Constraint work serialized over the batch limits throughput.} The engine calls
each request's grammar object in turn on its scheduler thread, between two forward passes,
which suits a C++ automaton that answers in microseconds. An exact check is heavier: opening
a field costs up to 10\,ms at the 99th percentile (\cref{sec:eval-enforce}). Run in a
single worker process beside the GPU, the check holds the masks to 0.90 of \xg on meeting
planning and our drafter to 0.92 to 0.93 of the engine's n-gram drafter; with eight worker
processes the masks reach 0.95 and the drafter 1.09 to 1.22
(\cref{fig:engines}(g)). The work must leave the scheduler thread and spread over processes
that run while the device computes (\cref{sec:service}).

\section{Design}
\label{sec:design}

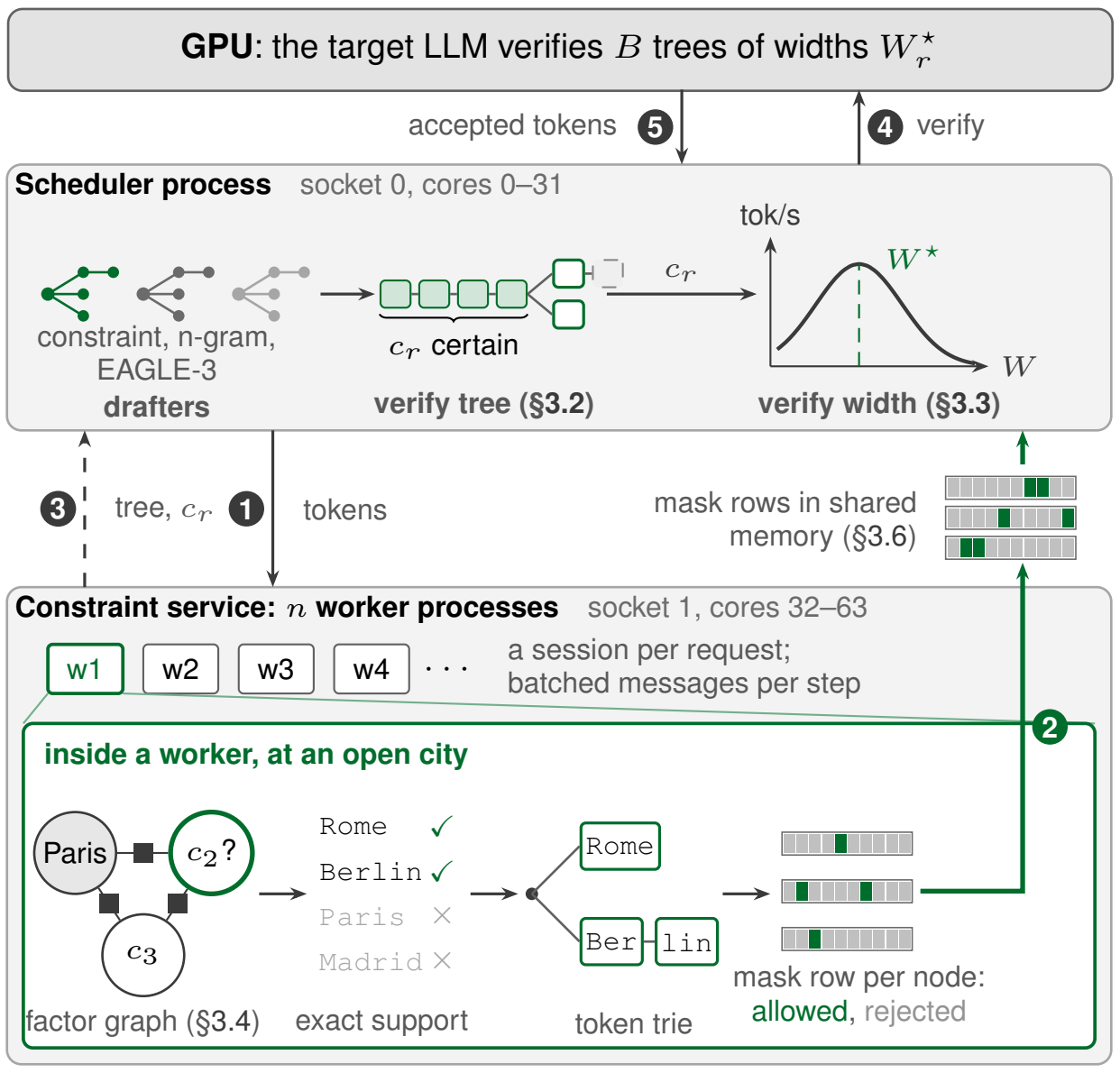
\begin{figure}[t]
\centering
{\sffamily\hypersetup{linkcolor=cLine}%
\begin{tikzpicture}[font=\scriptsize,
  region/.style={rounded corners=3pt, line width=0.5pt},
  rhead/.style={font=\tiny\bfseries, anchor=north west, inner sep=1.5pt},
  lbl/.style={font=\tiny, text=cLine!85, align=center},
  var/.style={circle, draw=cLine, line width=0.5pt, fill=white, minimum size=0.6cm,
    inner sep=0pt, font=\tiny},
  fac/.style={rectangle, fill=cLine, inner sep=0pt, minimum size=4pt},
  mini/.style={minimum height=0.2cm, minimum width=0.22cm, inner sep=0.8pt},
  arr/.style={->, line width=0.6pt, cLine},
  rarr/.style={->, line width=1pt, cCombined},
  stepno/.style={num, font=\tiny\bfseries, inner sep=0.7pt},
]
\newcommand{\maskbar}[3]{%
  \foreach \k in {0,...,9} \fill[cLine!32] (#1+\k*0.075,#2) rectangle ++(0.065,0.11);
  \foreach \k in {#3} \fill[cCombined] (#1+\k*0.075,#2) rectangle ++(0.065,0.11);
  \draw[cLine!75, line width=0.3pt] (#1-0.012,#2-0.012) rectangle (#1+0.752,#2+0.122);
}
\newcommand{\minitree}[3]{%
  \fill[#3] (#1,#2) circle (1.4pt);
  \foreach \dy in {0.16,0,-0.16} {
    \draw[#3, line width=0.5pt] (#1,#2) -- (#1+0.26,#2+\dy);
    \fill[#3] (#1+0.26,#2+\dy) circle (1.2pt);
  }
  \draw[#3, line width=0.5pt] (#1+0.26,#2+0.16) -- (#1+0.48,#2+0.16);
  \fill[#3] (#1+0.48,#2+0.16) circle (1.2pt);
}

\node[draw=cLine!80, fill=cLine!14, region, minimum width=8.1cm, minimum height=0.6cm,
  anchor=north west] (gpu) at (0,7.75)
  {\textbf{GPU}: the target LLM verifies $B$ trees of widths $W_r^\star$};

\draw[region, fill=cEngine, draw=cLine!45] (0,4.65) rectangle (8.1,6.6);
\node[rhead] at (0,6.6) {Scheduler process\quad\textmd{\textcolor{cLine!75}{socket 0, cores 0--31}}};
\minitree{0.3}{5.65}{cCombined}
\minitree{1.0}{5.65}{cLine!75}
\minitree{1.7}{5.65}{cLine!45}
\node[lbl] at (1.1,5.22) {constraint, n-gram,\\EAGLE-3};
\node[lbl, font=\tiny\bfseries] at (1.1,4.83) {drafters};
\node[det, mini] (v1) at (2.85,5.65) {};
\node[det, mini, right=0.05cm of v1] (v2) {};
\node[det, mini, right=0.05cm of v2] (v3) {};
\node[det, mini, right=0.05cm of v3] (v4) {};
\node[cand, mini, anchor=west] (v5) at ($(v4.east)+(0.18,0.15)$) {};
\node[cand, mini, anchor=west] (v6) at ($(v4.east)+(0.18,-0.15)$) {};
\node[graft, mini, right=0.05cm of v5] (v7) {};
\foreach \a/\b in {v1/v2,v2/v3,v3/v4} \draw[edge] (\a) -- (\b);
\draw[edge] (v4.east) -- (v5.west);
\draw[edge] (v4.east) -- (v6.west);
\draw[edge, dashed] (v5) -- (v7);
\draw[decorate, decoration={brace, amplitude=1.5pt, mirror}, line width=0.4pt]
  ($(v1.south west)+(0,-0.04)$) -- ($(v4.south east)+(0,-0.04)$)
  node[midway, below=0.5pt, font=\tiny] {$c_r$ certain};
\node[lbl, font=\tiny\bfseries] at (3.5,4.83) {verify tree (\S\ref{sec:proposer})};
\draw[arr] (2.3,5.65) -- (2.7,5.65);
\begin{scope}[shift={(5.6,5.12)}]
  \draw[->, line width=0.4pt, cLine] (0,0) -- (1.6,0) node[right=-1pt, font=\tiny] {$W$};
  \draw[->, line width=0.4pt, cLine] (0,0) -- (0,0.95) node[above=-1pt, font=\tiny] {tok/s};
  \draw[cLine, line width=0.8pt] plot[smooth, domain=0.05:1.5] (\x, {0.75*exp(-((\x-0.65)^2)/0.2)});
  \draw[cCombined, densely dashed, line width=0.4pt] (0.65,0) -- (0.65,0.75);
  \node[font=\tiny, text=cCombined, anchor=west] at (0.72,0.8) {$W^\star$};
\end{scope}
\node[lbl, font=\tiny\bfseries] at (6.4,4.83) {verify width (\S\ref{sec:width})};
\draw[arr] (4.4,5.65) -- (5.5,5.65) node[midway, above=-1pt, font=\tiny] {$c_r$};

\foreach \r in {0,...,2} {
  \pgfmathtruncatemacro{\first}{mod(\r*3+1,10)}
  \pgfmathtruncatemacro{\second}{mod(\r*7+2,10)}
  \begin{scope}[shift={(6.9,3.72+\r*0.22)}, scale=1.25] \maskbar{0}{0}{\first,\second} \end{scope}
}
\node[lbl, anchor=east, align=right] at (6.8,3.98) {mask rows in shared\\memory (\S\ref{sec:service})};

\draw[region, fill=cEngine, draw=cLine!45] (0,0) rectangle (8.1,3.5);
\node[rhead] at (0,3.5) {Constraint service: $n$ worker processes\quad\textmd{\textcolor{cLine!75}{socket 1, cores 32--63}}};
\foreach \i/\x in {1/0.3, 2/1.0, 3/1.7, 4/2.4} {
  \draw[fill=white, draw=cLine!80, rounded corners=1pt] (\x,2.72) rectangle ++(0.55,0.36);
  \node[font=\tiny] at (\x+0.275,2.9) {w\i};
}
\draw[fill=white, draw=cCombined, line width=0.7pt, rounded corners=1pt] (0.3,2.72) rectangle ++(0.55,0.36);
\node[font=\tiny, text=cCombined] at (0.575,2.9) {w1};
\node[font=\scriptsize] at (3.25,2.9) {$\cdots$};
\node[lbl, anchor=west, align=left] at (3.55,2.9) {a session per request;\\batched messages per step};
\draw[cCombined!55, line width=0.4pt] (0.3,2.72) -- (0.12,2.5);
\draw[cCombined!55, line width=0.4pt] (0.85,2.72) -- (7.98,2.5);
\draw[fill=white, draw=cCombined, line width=0.7pt, rounded corners=2pt] (0.12,0.12) rectangle (7.98,2.5);
\node[font=\tiny\bfseries, text=cCombined, anchor=north west] at (0.15,2.48) {inside a worker, at an open city};
\node[var, fill=cLine!12] (c1) at (0.5,1.55) {Paris};
\node[var, draw=cCombined, line width=1pt] (c2) at (1.5,1.55) {$c_2$?};
\node[var] (c3) at (1.0,0.8) {$c_3$};
\node[fac] at (1.0,1.55) {};
\node[fac] at (0.75,1.18) {};
\node[fac] at (1.25,1.18) {};
\draw[cLine, line width=0.4pt] (c1) -- (c2);
\draw[cLine, line width=0.4pt] (c1) -- (c3);
\draw[cLine, line width=0.4pt] (c2) -- (c3);
\node[lbl] at (1.0,0.3) {factor graph (\S\ref{sec:engine})};
\draw[arr] (1.85,1.25) -- (2.2,1.25);
\begin{scope}[every node/.style={font=\tiny\ttfamily, anchor=west, inner sep=1pt}]
  \node at (2.25,1.75) {Rome};   \node[text=cCombined, font=\tiny] at (3.05,1.75) {\checkmark};
  \node at (2.25,1.42) {Berlin}; \node[text=cCombined, font=\tiny] at (3.05,1.42) {\checkmark};
  \node[text=cLine!45] at (2.25,1.09) {Paris};  \node[text=cLine!45, font=\tiny] at (3.05,1.09) {$\times$};
  \node[text=cLine!45] at (2.25,0.76) {Madrid}; \node[text=cLine!45, font=\tiny] at (3.05,0.76) {$\times$};
\end{scope}
\node[lbl] at (2.75,0.3) {exact support};
\draw[arr] (3.4,1.25) -- (3.75,1.25);
\fill[cLine] (3.85,1.25) circle (1.3pt);
\node[cand, font=\tiny\ttfamily, inner sep=1pt, anchor=west] (r1) at (4.2,1.6) {Rome};
\node[cand, font=\tiny\ttfamily, inner sep=1pt, anchor=west] (r2) at (4.2,0.9) {Ber};
\node[cand, font=\tiny\ttfamily, inner sep=1pt, right=0.08cm of r2] (r3) {lin};
\draw[edge] (3.85,1.25) -- (r1.west);
\draw[edge] (3.85,1.25) -- (r2.west);
\draw[edge] (r2) -- (r3);
\node[lbl] at (4.6,0.3) {token trie};
\draw[arr] (5.25,1.25) -- (5.6,1.25);
\begin{scope}[shift={(5.7,1.55)}, scale=1.25] \maskbar{0}{0}{4} \end{scope}
\begin{scope}[shift={(5.7,1.2)}, scale=1.25] \maskbar{0}{0}{1,6} \end{scope}
\begin{scope}[shift={(5.7,0.85)}, scale=1.25] \maskbar{0}{0}{2} \end{scope}
\node[lbl] at (6.25,0.5) {mask row per node:\\\textcolor{cCombined}{allowed}, \textcolor{cLine!55}{rejected}};

\draw[arr] (1.95,4.65) -- (1.95,3.5) node[stepno, midway, left=1.5pt] {1};
\node[lbl, anchor=west] at (2.05,4.07) {tokens};
\draw[arr, dashed] (0.57,3.5) -- (0.57,4.65) node[stepno, midway, left=1.5pt] {3};
\node[lbl, anchor=west] at (0.67,4.07) {tree, $c_r$};
\draw[rarr] (6.7,1.27) -| (7.45,3.68) node[stepno, fill=cCombined, pos=0.75, right=1.5pt] {2};
\draw[rarr] (7.45,4.4) -- (7.45,4.65);
\draw[arr] (6.25,6.6) -- (6.25,7.15) node[stepno, midway, right=1.5pt] {4};
\node[lbl, anchor=west] at (6.55,6.875) {verify};
\draw[arr] (4.95,7.15) -- (4.95,6.6) node[stepno, midway, left=1.5pt] {5};
\node[lbl, anchor=east] at (4.6,6.875) {accepted tokens};
\end{tikzpicture}}
\caption[\sys in a serving engine]{\sys in a serving engine. \circled{1} The scheduler sends each request's accepted
tokens to its worker. \circled{2} The worker computes the open field's exact support, walks
its token trie, and writes the mask row of every draft node into shared memory.
\circled{3} It replies with the tree and its certain nodes $c_r$, which the target accepts
with probability one; the engine merges the tree with its drafters and picks each
request's width $W_r^\star$. \circled{4} The target verifies under the rows. \circled{5} The accepted tokens
start the next step. The worker panel shows an open position of another tour.}
\label{fig:overview}
\end{figure}

\Cref{fig:overview} shows \sys inside a serving engine. A request names a \emph{spec}, a
template with a factor graph, together with an instance of the problem. After the
draft--verify loop (\cref{sec:drafts}), the section follows \cref{sec:observations}: the
drafters and widths take O1's opportunity (\cref{sec:proposer,sec:width}), exact
feasibility and tokenization meet O2 (\cref{sec:engine,sec:tokens}), and the service meets
O3 (\cref{sec:service}).

\subsection{Lossless constraint drafts}
\label{sec:drafts}

\sys decodes in the draft--verify loop of speculative decoding, with the constraint as
drafter. At each step it drafts, for each request, the run of tokens the relations
determine from the current position, then candidates at the first open position, and the
target verifies the tree in one forward pass under a mask row for every node. A determined
token is the one token its mask admits, so the target accepts it with probability one, and
the constraint knows before the verify how many of each request's drafts are certain.

The drafts are lossless. Our reference decoder is token-by-token decoding under the mask
``grammar and relational feasibility''. Speculation changes only which tokens one forward
pass verifies, so in exact arithmetic greedy output equals the reference decoder's and
exact rejection sampling preserves its token-level distribution. That distribution is the
masked decoder's, the guarantee grammar masks
provide~\cite{park2024grammar}, which differs from the model's distribution conditioned on
a valid record. In floating point a
verify of another shape can flip a near-tie (\cref{sec:limits}); validity holds for every
record.

\textbf{A step, end to end.} Take the tour of \cref{fig:declare} with the road rows shown,
after the model has written the first stop, \code{Rome} on day 1, and the key of the
second stop's city. The scheduler sends the request's accepted tokens to its worker (\circled{1}). The worker adds
\code{Rome} and day 1 as evidence and asks the factor graph for the support of the second
city: the road table admits only \code{Oslo}, so that field and its day, $1+2=3$, are
determined, and so is the third stop, \code{Lima} on day 6. The rest of the record is a
determined run of 24 Qwen3 tokens in the engine's JSON rendering. The worker writes a mask row for each node of the run,
up to the node budget, into the request's shared-memory block (\circled{2}) and replies that
every node is certain (\circled{3}). The scheduler
gives the request the server's full width of 12 slots, the target verifies them under the
rows (\circled{4}), and all 12 are accepted (\circled{5}). Meanwhile the worker has already computed the remaining 12 tokens, so the next step reads
them and completes the record at once.

\subsection{The drafters}
\label{sec:proposer}

O1 shows that the relations determine most of each record. \sys therefore drafts from the
constraint, as \cref{sec:drafts} describes, and combines that draft with the engine's
drafters where the record holds text that the relations leave undetermined.

\textbf{Factor drafter.} The factor drafter proposes the constraint's tree
(\cref{fig:overview}, left): the determined run from the current position, then the
candidates of the token trie at the first open position (\cref{sec:tokens}), up to the
node budget. The target accepts every node of the determined run, because the mask at
each of those positions allows one token.

\textbf{Combined drafter.} At an open position the model chooses among the candidates, and
on copy workloads the engine's n-gram corpus often predicts its choice. The combined
drafter grafts the corpus's continuation under each open leaf and asks the service for the
grafted nodes' rows. Rows go to the grafted tokens the spec allows, and the mask walk
descends through allowed tokens alone, so a refused graft drops out of the verify.

\textbf{Merged drafter.} Where the open text is prose, a learned head may predict it better
than a corpus. The merged drafter therefore gives a request the constraint's tree whenever
the constraint offers one and any other request EAGLE-3's, in one verify at one width per
step, and skips the draft head when every request takes the constraint's tree.

The engine's drafters also run under our masks, which the evaluation uses to compare
drafters on identical records; the service writes their rows while the target runs.

\subsection{Per-request verify widths}
\label{sec:width}

The drafts also tell the engine how wide to verify. A verify step sends $T$ tokens through
the target. Below the device's knee, where reading the weights bounds the forward pass,
extra slots are nearly free; above it, each slot costs time and returns a token when the
target accepts it. An engine verifies every request of a step at one width, so a request
inside free text takes the slots a request inside a determined run needs. The constraint
knows, per request, how many of its draft positions are certain, so the factor and
combined drafters give each request $r$ a separate width $W_r$ and choose the widths to
maximize expected accepted tokens per unit of step time,
\begin{equation}
\label{eq:width}
\{W_r^\star\} = \arg\max_{\{W_r\}} \frac{\sum_{r} \bigl(1 + c_r + g(W_r - 1 - c_r)\bigr)}
  {a + b \cdot \max\bigl(K, F(\textstyle\sum_r W_r)\bigr)},
\end{equation}
where $c_r = c_r(W_r)$ counts request $r$'s certain nodes within its width, which the
target accepts with probability one; $g(k)$ is the expected accepted tokens of $k$ guessed
nodes, an exponential moving average over the drafter's verifies, each of which measures
every narrower width at once; and $F$ is the tokens the forward pass computes for the step's slots,
padded to a captured graph when one runs the step. Read through its least concave
majorant, a request's return per added slot is non-increasing, so for each total the best widths
take the slots of highest return across requests, and one sorted pass scans every total
(\cref{alg:width} in \cref{app:design}). Certain nodes return one token each and are taken
first; a request inside free text keeps its root and the guesses whose return exceeds their
cost.

The knee $K$ is 128 tokens, measured on the A6000; $b$ is fitted once per model and
tensor-parallel
size; and $a$ is learned from the drafter's measured steps, pooled across ranks at
intervals so that every rank chooses the same widths. Per-request widths are a server mode,
on in every run we report except the full sets, the worker ablation and the jump-forward
comparison, which use one width per step.

The forward pass computes each request's real slots alone; \cref{app:design} describes
the verify layout, the captured graphs and how tensor-parallel ranks agree on widths.

\subsection{Exact feasibility}
\label{sec:engine}

\begin{figure}[t]
\scriptsize
\begin{verbatim}
{"schema": {"type": "object", "properties": {"stops":
  {"type": "array", "minItems": 3, "maxItems": 3,
   "items": {"type": "object", "properties": {
    "city": {"enum": ["Rome", "Oslo", "Lima"]},
    "day": {"type": "integer", "minimum": 1,
            "maximum": 6}}}}}},
 "relations": [
  {"kind": "allowed_rows",
   "paths": ["/stops/*/city", "/stops/*+1/city"],
   "rows": [["Rome", "Oslo"], ["Oslo", "Lima"]]},
  {"kind": "all_different", "path": "/stops/*/city"},
  {"kind": "running_sum", "over": "/stops/*/city",
   "target": "/stops/*/day", "base": 1,
   "weights": [["Rome",2], ["Oslo",3], ["Lima",1]]}]}
\end{verbatim}
\vspace{-2ex}
\caption{A declared record: a three-city tour that takes a road between stops, visits each
city once and reaches each stop on day one plus the stays before it. Abridged from our
repository's example: required keys, other declaration keys, labels, one road row and a
first-day relation omitted.}
\label{fig:declare}
\end{figure}

O2 requires every draft, and every token the target samples, to hold the relations exactly.

\textbf{Specs.} A client declares a record as a JSON Schema and a list of relations over
JSON Pointer paths (\cref{fig:declare}), and the server compiles it to a \emph{spec}: a
template, whose choice fields are factor variables, and a factor graph over those
variables. A spec is built from the request alone: the variables' domains and the factors
come from the problem statement or the tool context in the prompt. A spec can be built at
three levels, which the evaluation compares:
\emph{syntax}, which a grammar engine can enforce; \emph{membership}, where each reference
is one of the identifiers the context declares; and \emph{relational}, with every factor.

\textbf{Exact feasibility.} Every mask rests on one query: the support
(\cref{sec:factorgraph}) of the next field given the values the decoder has committed. To
answer it, the factor engine stores each factor as a dense Boolean table within a per-graph
budget and refuses an input table over the budget when the spec is built. A support query
reduces those tables by the evidence and eliminates the other variables in a min-fill order
in the Boolean semiring: the tables of a bucket are joined by AND and the eliminated
variable is reduced by OR. The answer is exact: every value in the support has a valid
completion.

A decoder's queries form a walk, each clamping one more variable, so consecutive queries
share almost all of their work. The engine exploits this by memoizing a reduced table by
its factor and the restriction the evidence places on its scope, and a bucket's message by
its variable and the keys of the tables that feed it. A query then reduces only the tables
the new value changes and joins only the buckets they reach; on a chain-shaped graph the
cost is the work of one neighborhood.

Memoization bounds the work per query, but the size of each table depends on how a relation
is encoded. Trip planning's all-different over
cities, encoded as one table over every city variable, joins up to 22.2 million entries at
seven cities and exceeds the budget at eight. A chain encoding over visited-set states
holds the same itineraries with a largest join of 8,575 entries at seven cities and 529,200
at ten, which covers every problem of the dataset.

\textbf{Global factors.} Some relations stay large under any table encoding, yet are cheap to
decide directly. A meeting
schedule visits parties in turn, each with a location, an availability window and a
minimum stay; as tables it is a join over three clock variables per visit of several
hundred values each. Because waiting is allowed, ending a visit earlier keeps every way to
continue, so whether a plan can still meet $r$ more parties depends only on the parties
met, the last one and the time, monotonically in the time. A dynamic program over
$2^P \cdot P$ states, the parties met and the last party, therefore gives the latest time
from which each state can still complete the plan, and every field's exact support is a
lookup in it. Such a relation enters the graph as a \emph{global factor} with a dedicated
exact check, over a scope disjoint from every other factor's, so a variable's support is the
support within its component while the rest remain satisfiable.

\subsection{From values to tokens}
\label{sec:tokens}

The feasibility check answers questions about values, but the decoder emits tokens. The
structure engine bridges the two and maintains the invariant that every admitted token
extends to a complete, valid record. Outside free text, a state holds candidate
continuations: each runs from the current position through the next field value or
structural decision and on through the literal text that follows, up to the next field
boundary. These candidates are tokenized in context, as the tokens they add to the
tokenization of the preceding text, so a value's first token is the one the tokenizer
produces after the preceding text. The allowed set is then the next token of every candidate
that matches the tokens emitted so far. A value followed by a decision, such as whether an
array goes on, gives one candidate per pair, because the token that closes the value differs
between the two.

A free field is a token class that ends at the first token of the following
literal; where the tokenizer merges punctuation with a delimiter, as in \code{!"}, the
engine splits the merged token and carries the rest of the delimiter as a forced run. If a
boundary would re-tokenize the text before it, the request ends with an explicit error.
States are memoized per parent and token, so a draft tree and the later walk over it derive
each state once. Positions where one token is allowed form the \emph{determined runs} that
the drafters of \cref{sec:proposer} propose.

\textbf{The session.} A per-request session exposes \code{fill}, \code{accept},
\code{rollback}, \code{fork} and \code{expand}, which builds a draft tree under a node
budget. It keeps the state after every accepted token, so a rollback is a list truncation
and a fork shares the states it has; a disallowed token or an empty allowed set raises an
error before it reaches the sampler. Every drafted and verified token passes through the
session, which is why the guarantee of \cref{sec:drafts} holds.

\subsection{The constraint service}
\label{sec:service}

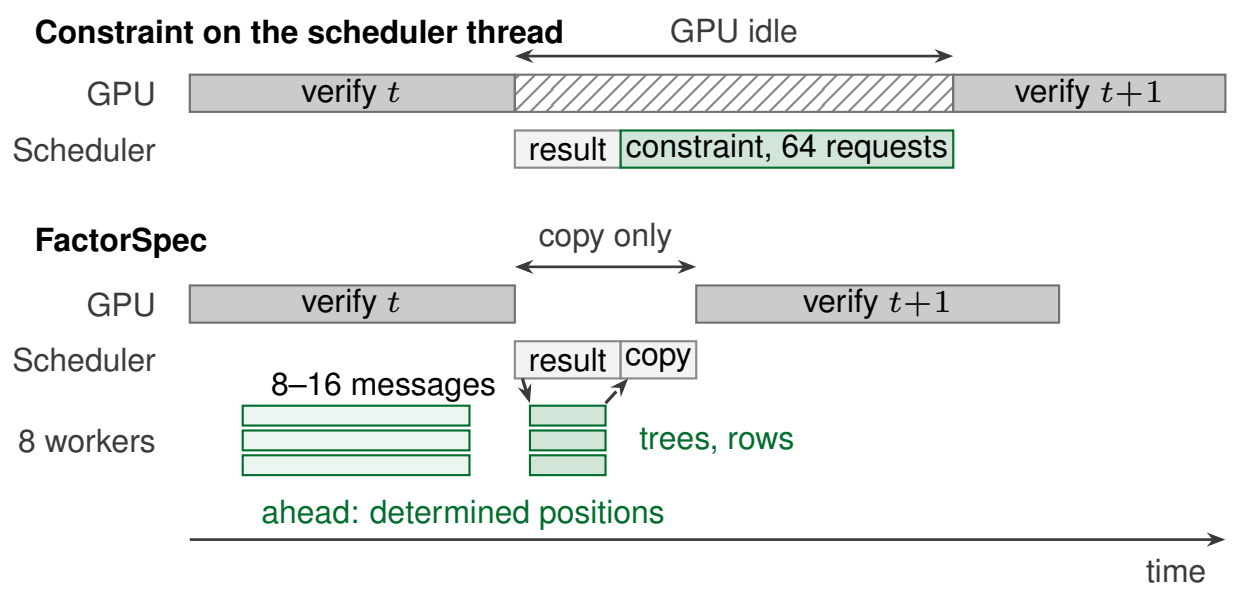
\begin{figure}[t]
\centering
{\sffamily
\begin{tikzpicture}[font=\tiny, yscale=0.82,
  lanel/.style={font=\tiny, text=cLine, anchor=east},
  case/.style={font=\tiny\bfseries, anchor=west},
  msg/.style={->, line width=0.6pt, cLine},
  gapmark/.style={<->, line width=0.5pt, cLine},
]
\node[case] at (0,4.75) {Constraint on the scheduler thread};
\node[lanel] at (1.05,4.25) {GPU};
\node[lanel] at (1.05,3.8) {Scheduler};
\draw[gpu] (1.15,4.1) rectangle (3.3,4.4);     \node at (2.22,4.25) {verify $t$};
\draw[idle] (3.3,4.1) rectangle (6.2,4.4);
\draw[gpu] (6.2,4.1) rectangle (8.0,4.4);      \node at (7.1,4.25) {verify $t{+}1$};
\draw[cpu] (3.3,3.65) rectangle (4.0,3.95);    \node at (3.65,3.8) {result};
\draw[work] (4.0,3.65) rectangle (6.2,3.95);   \node at (5.1,3.8) {constraint, 64 requests};
\draw[gapmark] (3.3,4.55) -- (6.2,4.55) node[midway, above=-1pt, text=cLine] {GPU idle};

\node[case] at (0,3.05) {\sys};
\node[lanel] at (1.05,2.55) {GPU};
\node[lanel] at (1.05,2.1) {Scheduler};
\node[lanel] at (1.05,1.45) {8 workers};
\draw[gpu] (1.15,2.4) rectangle (3.3,2.7);     \node at (2.22,2.55) {verify $t$};
\draw[gpu] (4.5,2.4) rectangle (6.9,2.7);      \node at (5.7,2.55) {verify $t{+}1$};
\draw[cpu] (3.3,1.95) rectangle (4.0,2.25);    \node at (3.65,2.1) {result};
\draw[cpu] (4.0,1.95) rectangle (4.5,2.25);    \node at (4.25,2.1) {copy};
\foreach \y in {1.25,1.45,1.65} {
  \draw[work, fill=cCombined!7] (1.5,\y-0.08) rectangle (3.0,\y+0.08);
  \draw[work] (3.4,\y-0.08) rectangle (3.9,\y+0.08);
}
\node[anchor=north west, text=cCombined] at (1.5,1.12) {ahead: determined positions};
\node[anchor=west, text=cCombined] at (4.0,1.45) {trees, rows};
\draw[msg] (3.35,1.95) -- (3.4,1.75);
\node[anchor=east] at (3.3,1.88) {8--16 messages};
\draw[msg, dashed] (3.9,1.75) -- (4.05,1.95);
\draw[gapmark] (3.3,2.85) -- (4.5,2.85) node[midway, above=-1pt, text=cLine] {copy only};
\draw[->, line width=0.5pt, cLine] (1.15,0.65) -- (8.0,0.65) node[anchor=north east] {time};
\end{tikzpicture}}
\caption{One decode step at 64 requests in flight, schematic. On the scheduler thread the
GPU waits for the constraint's work; with the service, workers compute ahead during verify
$t$ and build the trees while the scheduler processes the result.}
\label{fig:timeline}
\end{figure}

O3 requires the constraint's work to leave the scheduler thread (\cref{fig:timeline}). The
service runs it in a
pool of worker processes, eight per rank by default, on the cores the scheduler process
leaves free; a worker imports the constraint library alone, and every session of one spec opens
on one worker and shares its compiled check.

Each request holds two blocks of a shared-memory array of mask rows, sized to the verify
width, one for the current step and one for the next. A worker writes each node's row into
a block and its reply names the block, so the scheduler's \code{fill} is a row copy. Per
step, each request posts its accepted tokens and its ask: the constraint's tree, the rows of
another drafter's tree, or the root row. The messages leave in one batch per worker and
kind of ask, so a step of 64 requests costs 8 to 16 messages each way; \cref{alg:worker} in
\cref{app:design} gives a worker's step.

At a determined position the next token is known before the target samples it, so after
replying the worker computes the next reply ahead while the device runs the verify, one
field value at a time, so that the next step finds the worker ready even while it opens a
field of several hundred values. A speculative drafter asks for the next trees as soon as
the verify's tokens reach the host, before the scheduler accepts them into the grammars.
The workers therefore build the trees while the scheduler thread processes the result, and
the scheduler copies rows before it launches the next verify (\cref{fig:timeline}).

\subsection{Engine integration}
\label{sec:integration}

\begin{figure*}[t]
\begin{minipage}[b]{0.665\textwidth}
\centering
\includegraphics{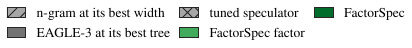}\\
\subfloat[8 in flight]{\includegraphics{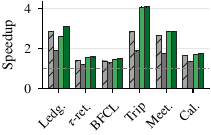}}%
\subfloat[64 in flight]{\includegraphics{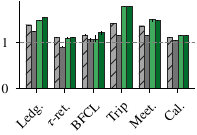}}%
\subfloat[128 in flight]{\includegraphics{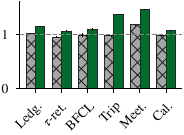}}
\end{minipage}\hfill
\begin{minipage}[b]{0.32\textwidth}
\centering
\includegraphics{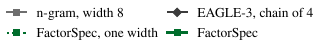}\\
\includegraphics{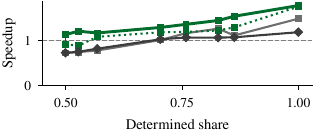}
\end{minipage}\\
\begin{minipage}[t]{0.665\textwidth}
\caption{With every record valid, the combined drafter matches or outruns the speculator
tuned per load. Throughput over \xg alone, each speculator at its best shape per cell.}
\label{fig:spec}
\end{minipage}\hfill
\begin{minipage}[t]{0.32\textwidth}
\caption{As notes grow, only the combined drafter stays above \xg. Trip with notes at 64 in
flight.}
\label{fig:sweep}
\end{minipage}
\end{figure*}

\begin{figure*}[t]
\centering
\includegraphics{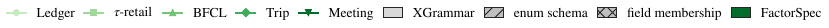}\\
\subfloat[Masks over \xg]{\includegraphics{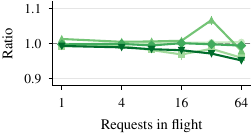}}%
\subfloat[Valid by constraint]{\includegraphics{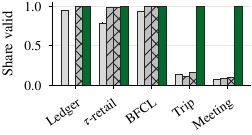}}%
\subfloat[Arc consistency]{\includegraphics{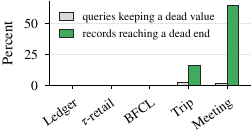}}%
\subfloat[Workers on the host]{\includegraphics{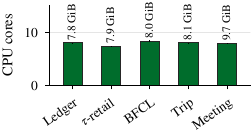}}
\caption{Exactness costs what a grammar mask costs, and every record completes. (b) at 8 in
flight, enum schemas on every workload except the ledger; (c) over 200 gold records and records served at 64 in flight; (d) at 64 in flight.}
\label{fig:breakdown}
\end{figure*}

\textbf{SGLang.} \sys registers a grammar backend and five speculative algorithms through
SGLang's plugin registries: the three drafters, and the engine's n-gram and EAGLE-3
drafters under \sys's masks. The backend hands every non-relational request to
\xg, so the server serves everything an \xg server serves. Every configuration we evaluate,
ours included, runs under SGLang's overlap rule, which disables the overlap scheduler
for a speculative batch with a grammar.

\textbf{vLLM.} vLLM drives the grammar one request at a time, and a blocking call per
request would make the constraint's work serial again. The vLLM grammar object therefore
answers \code{fill}, \code{accept} and draft validation from the block of its last reply,
and the proposer posts every request's step before it waits for any reply.

\section{Implementation}
\label{sec:implementation}

\sys is 16,700 lines of Python, counted with \code{wc}. The constraint engine, 9,100 lines
excluding its workload declarations, holds the factor engine, the structure engine, the
session and the JSON Schema frontend, and runs independently of any serving engine; the
serving layer, 7,600 lines, holds the service, the drafters, the
width model and both engines' adapters. A declaration's schema holds objects, enums,
integer ranges, optional properties, bounded arrays and capped free text, and its relations
are drawn from twelve kinds, from allowed rows and all-different to running sums and visit
schedules, whose encodings a compiler chooses; a declaration compiles in 0.05 to 0.55\,ms
at the median. In SGLang 0.5.16, \sys registers its grammar backend and drafters through
the plugin registries and replaces three functions that lie outside them: the batch mask
fill and the draft-tree mask builders of the n-gram and EAGLE-3 workers;
in vLLM 0.21.0, two patches register the backend and the proposer. The workers are plain
subprocesses that share the mask rows through one shared-memory segment. A conformance test replays recorded traces through both engines' grammar objects and
compares every mask row against the constraint session's output.

\section{Evaluation}
\label{sec:eval}

We ask four questions. \textbf{RQ1:} How much faster does the constraint make decoding as a
drafter (\cref{sec:eval-proposer})? \textbf{RQ2:} What does exactness cost against a
grammar mask (\cref{sec:eval-enforce})? \textbf{RQ3:} How many more valid records does an
operator get (\cref{sec:eval-goodput})? \textbf{RQ4:} Does the design generalize, and what
accounts for it (\cref{sec:eval-engines})?

\subsection{Setup}
\label{sec:eval-setup}

\textbf{Platform.} Two NVIDIA RTX A6000 GPUs and a 64-core two-socket host; Qwen3-8B with
an EAGLE-3 head unless stated, and Llama 3.1 8B, Qwen3-1.7B and Qwen3-14B; SGLang 0.5.16
and vLLM 0.21.0 (\cref{app:setup} gives the details).

\textbf{Workloads.} Three tool-calling workloads, a synthetic \emph{ledger} and
\emph{$\tau$-bench retail}~\cite{yao2024tau} and \emph{BFCL
multi-turn}~\cite{patil2025berkeley} calls, whose arguments must name identifiers the
context declares; and three NATURAL PLAN~\cite{zheng2024natural} planning workloads,
\emph{trip}, \emph{meeting} and \emph{calendar}, of 999 problems each, plus all 1,599 trip
problems as a full set. In meeting and calendar planning the model also chooses among
valid plans. \emph{Trip planning with notes} adds a free-text note of up to $N$ tokens to
every stay, which lowers the determined share from 1.00 at $N{=}0$ to 0.50 at 128. RQ2 and
RQ4 use the first five workloads, and RQ2 all six at 128 in flight. \Cref{app:setup}
states each workload's relations and prompts.

\textbf{Baselines.} Constraints: \xg and llguidance on the syntax schema, and \xg with
each field's value set as an \emph{enum}. Speculators: SGLang's n-gram drafter at its
default width of 12 and at 4, 8, 16, 24 and 32, and EAGLE-3 at its default tree, at five
steps four ways and as chains of two, four and eight tokens. We call each of these eleven
drafter settings a \emph{shape}, one workload at one load a \emph{cell}, and a cell's
fastest shape its \emph{tuned speculator}. Retries: \emph{retry} samples up to three attempts, \emph{error feedback} names the
violated relation, and \emph{thinking} lets Qwen3 reason for up to 8,192 tokens. Solvers:
\emph{solve then emit} computes a valid record on the CPU before decoding and constrains \xg to emit it;
an \emph{enumerated schema} lists each request's feasible records as a JSON Schema
\code{anyOf} of at most 1,024 records. We report \sys's combined drafter unless a figure names another; all configurations run
under SGLang's overlap rule.

\textbf{Metrics and protocol.} We report throughput, and \emph{valid records per
GPU-second}, which charges every invalid answer and retry; \emph{correct records per
GPU-second} counts only records that match gold. Closed loop runs at 1 to 128 requests in
flight on a separate window of records per load. Open loop offers Poisson arrivals at six
rates for two minutes each; the planning workloads replay a pool of 128 problems, which
lets every n-gram corpus, ours included, see earlier answers (\cref{app:setup}). Each configuration runs on
a fresh engine, greedy except under retry and sampling; each of 332 configurations ran
three times, and figures show the mean and the range on one A6000 with Qwen3-8B unless they
name another setting. Validators written independently of the constraint engine check every
record.

\subsection{RQ1: The constraint as a drafter}
\label{sec:eval-proposer}

\begin{table}[t]
\centering
\caption{Tokens per verify step per request, the mean acceptance length, at 8 / 64 in
flight: each engine speculator at its fastest shape per cell, and our combined drafter.
Same runs as \cref{fig:spec}.}
\label{tab:accept}
\footnotesize
\setlength{\tabcolsep}{5pt}
\begin{tabular}{@{}lrrr@{}}
\toprule
Workload & n-gram & EAGLE-3 & \sys \\
\midrule
Ledger & 5.1 / 4.5 & 3.6 / 2.7 & 5.8 / 4.5 \\
$\tau$-bench retail & 5.2 / 4.9 & 2.5 / 2.5 & 6.4 / 5.1 \\
BFCL multi-turn & 4.0 / 3.2 & 2.8 / 2.4 & 4.3 / 4.1 \\
Trip planning & 5.9 / 5.3 & 3.9 / 2.6 & 10.5 / 9.9 \\
Meeting planning & 5.7 / 4.7 & 3.6 / 2.6 & 8.4 / 7.4 \\
Calendar planning & 6.8 / 3.6 & 4.1 / 3.1 & 6.9 / 6.5 \\
\bottomrule
\end{tabular}
\end{table}

\begin{figure}[t]
\centering
\includegraphics{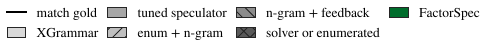}\\
\includegraphics{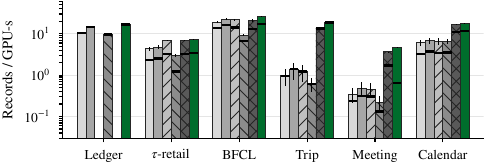}
\caption{\sys serves the most valid records per GPU-second (bars, log scale) on every
workload; on meeting planning the solver serves more that match gold (marks). At 64 in
flight, every invalid record and retry charged; four of the seven syntax-only baselines
shown; enum and solver baselines on all workloads but the ledger; 95\% intervals over
prompts and runs.}
\label{fig:valid}
\end{figure}

\begin{figure*}[t]
\centering
\includegraphics{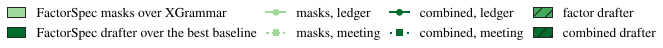}\\
\subfloat[vLLM]{\includegraphics{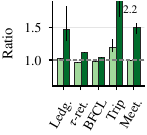}}%
\subfloat[8B, TP2]{\includegraphics{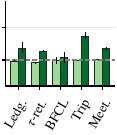}}%
\subfloat[Llama]{\includegraphics{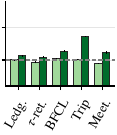}}%
\subfloat[1.7B]{\includegraphics{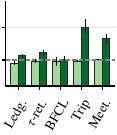}}%
\subfloat[14B, TP2]{\includegraphics{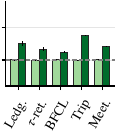}}%
\subfloat[Sampled]{\includegraphics{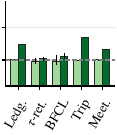}}%
\subfloat[Workers]{\includegraphics{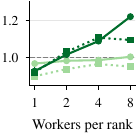}}%
\subfloat[Widths]{\includegraphics{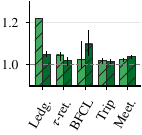}}
\caption{The drafters lead or tie on every workload in every setting, and the masks run at
0.95 to 1.19 of \xg. At 64 in flight; TP2 is two-GPU tensor parallelism; (f) is Qwen3-8B at
temperature one; (g) masks over \xg, combined over n-gram; (h) per-request versus fixed width.}
\label{fig:engines}
\end{figure*}

Following O1, we expect the constraint's drafts to gain most where the relations compute
the record, and to keep their gain at load. \Cref{fig:spec} compares the combined
drafter with the engine's speculators, each tuned over the eleven shapes per cell
(\cref{tab:shapes}).

\textbf{Faster than tuned speculators at every load.} With every record valid, the combined
drafter leads the tuned speculator by up to 1.43$\times$ at 8 in flight, 1.26$\times$ at
64 and 1.39$\times$ at 128. The lead is largest where the relations compute the record,
1.26 to 1.43$\times$ on trip planning, because there a determined run fills most of the
verify step. It is smallest where the model chooses most of the record and the n-gram corpus
covers the rest, 1.01 to 1.12$\times$ on the tool calls (a tie on $\tau$-bench retail at
64). At 128 in flight the speculators fall to 0.94 to 1.17$\times$ \xg, because a step
verifies every request at one width; the combined drafter still runs at 1.05 to
1.45$\times$, because per-request widths keep slots on the requests that return them.

\textbf{Longer accepted runs.} \Cref{tab:accept} shows why the throughput leads hold. Each
verify step accepts up to 10.5 tokens per request on trip planning and 8.4 on meeting
planning, 1.0 to 1.9$\times$ as many as the fastest n-gram shape, because a determined run
is accepted whole. Those determined runs carry most of the gain: verifying only them
reaches 0.89 to 0.97 of the factor drafter (RQ4). The guesses and per-request widths add
the rest, which decides the speed as free text grows: on the ledger at 64 in flight both
drafters yield 4.5 tokens per step, yet ours runs 1.11$\times$ faster, because per-request
widths spend fewer slots on guesses.

\textbf{The speed comes from the drafts.} A faster drafter could write shorter or different
records. Under our masks, where both drafters write identical records, the combined drafter
still runs at 1.07 to 1.42$\times$ the n-gram drafter at 8 in flight and 1.07 to
1.79$\times$ over loads 1 to 64.

\textbf{Free text.} As the determined share falls, so does the lead
(\cref{fig:sweep}). On trip planning with notes at 64 in flight, the combined drafter
falls from 1.77$\times$ \xg to 1.13$\times$ as notes grow to 128 tokens, while the tuned
n-gram drafter and EAGLE-3 fall below \xg from 32-token notes on, to 0.72 and 0.73.
Per-request widths explain the gap: a request inside a note keeps only the guesses whose
return exceeds their cost, which adds 8 to 37\% at every nonzero note budget. Removing
them drops our drafter to 0.88. The merged drafter, at one width per step, runs at 0.64 to
0.91 of the combined one.

\takeaway{The relations determine the drafts that lie beyond a speculator. Drafting them at
a width per request keeps speculation profitable at high load and in free text.}

\subsection{RQ2: The cost of exactness}
\label{sec:eval-enforce}

With the speed established, we ask what exactness costs.

\textbf{Masks at a grammar mask's cost.} The masks run at 0.95 to 1.07 of \xg's throughput
at every load up to 64 in flight (\cref{fig:breakdown}(a)) and at 0.94 to 1.13 at 128,
because the service computes each row while the device runs. Llguidance reads 0.94 to 1.05
on the same cells, so ratios above one come from the different records the masks produce,
which confirms that the check itself hides behind the step. That parity has a host cost:
the eight workers per rank hold 7.4 to 8.3 of the 32 cores of the socket the scheduler
leaves free and 7.8 to 9.7\,GiB (\cref{fig:breakdown}(d)). On Qwen3-1.7B, whose step is
under half as long, the masks still run at 0.95 to 1.01 of \xg.

\textbf{What one query costs.} The per-query breakdown explains where those cores go
(\cref{tab:querycost}). Most of each position's mask is a 4.3 to 4.9\,$\mu$s support
lookup, but opening a field rebuilds the elimination tables: that costs up to 2.2\,ms at
the 99th percentile on the tool calls and 10\,ms on meeting planning, where the dynamic
program is built. Serial on the scheduler thread, such openings would delay every request
of a step. Once the field is open, a single feasibility query into it takes 2 to 131\,$\mu$s
at the median and at most 3.2\,ms, and on trip planning the median grows from 101\,$\mu$s at
three cities to 198\,$\mu$s at ten as the tables grow. The memos make distinct queries 1.5
to 2.5$\times$ cheaper where the relations are tables, and change little on meeting and
calendar planning, whose global factors answer from dedicated checks.

\textbf{Weaker checks miss the cycles.} Exactness could be traded for speed: a weaker
constraint is cheaper to check, but it enforces only local consistency. An enum
schema or our field membership makes the tool calls valid, 0.98 to 1.00, because their
relations are mostly memberships. On trip and meeting planning, whose relations close
cycles, the weaker checks leave 0.08 to 0.16 valid (\cref{fig:breakdown}(b)). Our
arc-consistency propagator, memoized like the exact engine, is exact on the acyclic
relations of the tool
calls~\cite{freuder1982sufficient,beeri1983desirability}, at 0.59 to 0.97 of our masks'
throughput. On planning it runs at 0.29 to 0.95, and leads 12 to 16\% of trip records and
53 to 64\% of meeting records to infeasible prefixes (\cref{fig:breakdown}(c)), because
propagation sees each constraint in isolation; under the exact check, every record completes.
\label{sec:eval-exact}

\takeaway{Off the scheduler thread, the exact check's millisecond tail overlaps the GPU, so
exactness costs a grammar mask's time and about eight host cores per GPU. A weaker check
leads planning records to infeasible prefixes.}

\subsection{RQ3: Valid records}
\label{sec:eval-goodput}

An operator pays for valid records. Against a syntax-only baseline, we expect \sys to lead
by about the inverse of that baseline's validity; against
one that makes the records valid too, by throughput.

\textbf{Against the model alone.} The first expectation holds (\cref{fig:valid}). Against
the strongest of seven syntax-only baselines, \sys serves 2.6 to
13.9$\times$ the valid planning records per GPU-second at 8 and 64 in flight. The gap is
largest on trip planning, at least 9.3$\times$ over every prompt and run, because there the
baselines make fewer than one record in ten valid. On the tool calls, which an enum schema
already makes valid, the lead is throughput alone, 1.07 to 1.20$\times$. The lead carries
over to records that match gold, 2.0 to 13.8$\times$ on planning, with a tie on retail and
BFCL. Reasoning closes little of the gap: with thinking at 8 in flight on 64 problems, 19\% of
trip plans, 31\% of meeting schedules and 95\% of calendar slots become valid, so \sys
serves 79 to 631$\times$ their valid planning records per GPU-second.

\textbf{Against solvers and enumerated schemas.} The second expectation holds too. Solve
then emit needs only 0.3 to 1.7\,ms to compute each answer, yet it still enters the grammar
engine one token per step; \sys's determined runs write every known token at once, serving
1.28 to 1.46$\times$ the solver's valid itineraries and meeting schedules. The
enumerated schema fits every calendar and retail record and 381 of 384 BFCL calls, where
\sys leads it by 1.06 to 1.31$\times$; our enumeration, one schema per array item, refuses
every trip and meeting request, which an \code{anyOf} over whole records could list where
they are few. The solver chooses better, however, where gold follows a fixed objective: its
earliest meeting plans match gold in 46 to 56\% of problems against \sys's 14 to 16\%, so
it serves 2.6 to 2.8$\times$ \sys's correct meeting schedules, because its records always
match its objective.

\textbf{Full sets and load.} Over each full set (\cref{fig:full}), \sys's records are all
valid and match gold on 98 to 99\% of itineraries, 15.0\% of meeting schedules and 67.4\%
of calendar slots, against at most 13.6, 8.3 and 36.3\% for the baselines. Seen from the
request's side, just 9 to 14\% of the baselines' trip and meeting requests ever receive a
valid record; \sys's meeting schedules take 12.4\,s at the median, since the baselines' few
valid ones are short.

Under open-loop arrivals the gap widens further
(\cref{fig:time,fig:time-more}): past the grammar engine's capacity, \sys completes 32
valid itineraries per second inside 10\,s at 32 requests per second against 1.2 for the
n-gram drafter, and 7.2 valid meeting schedules inside 20\,s at 8 per second against 0.66.
Its 99th-percentile latency is the lowest at every rate, 3.8\,s against 26\,s on trip
planning, because the determined runs keep the server ahead.

\takeaway{Where relations join fields, validity, more than speed, sets how many usable
records a server delivers. Against an equally valid solver the margin is speed; where gold
follows a fixed objective, the solver chooses better.}

\subsection{RQ4: Generality and attribution}
\label{sec:eval-engines}
\label{sec:eval-ablation}

\textbf{Other engines, models and sampling.} The design carries over (\cref{fig:engines},
64 in flight). In vLLM, which drives the grammar one request at a time and verifies chains,
the factor drafter leads the best of six baselines by 1.03 to 2.24$\times$, because the
determined runs help even in chains. The masks reach 1.19$\times$ \xg on trip planning,
where the exact check's different records account for the gap. Across four models (Llama
3.1 8B, Qwen3-1.7B, Qwen3-8B, and Qwen3-14B over two GPUs), the combined drafter leads by
1.02 to 1.50$\times$, with ties on BFCL at 1.7B and at 8B over two GPUs. At temperature
one
validity stays 1.00 and the drafter leads by 1.02 to 1.35$\times$, because sampling adds
variety that the constraint still covers. Settings (b), (d), (e) and (f) ran at a later
commit (\cref{app:setup}).

\textbf{What each mechanism adds.} The three mechanisms each contribute measurable speed
when the others are in place (\cref{fig:engines}(g,\,h)). Eight workers per rank raise the
combined drafter by 19 to 31\% over one, as O3 anticipates: the check's millisecond tail
needs parallelism to hide. Jump-forward, which verifies only the determined run, reaches
0.89 to 0.97 of the factor drafter, so the guessed nodes return 4 to 12\% (0.94 to 1.01 in
a separate campaign, \cref{tab:wide}). Per-request widths add 2 to 22\% to the factor
drafter and 1 to 10\% to the combined drafter at 64 in flight, and cost up to 2.3\% at one
request in flight; one width per step, chosen by \cref{eq:width}, comes within 2\% of the
best fixed width.

\takeaway{The gains carry over to another engine, other models and sampling, and each
mechanism adds measurable speed when the others are in place.}

\section{Discussion and Limitations}
\label{sec:limits}
\enlargethispage{3\baselineskip}

A cell's three runs repeat its prompts, so their range measures timing, and a speed lead of
a few percent is a tie. Verify widths round differently: 1 to 4 of 64 greedy meeting records
differ, all valid. The host work hides behind the device's step, which we
measure down to Qwen3-1.7B's, on one GPU type and up to 128 requests in flight; a faster
GPU shortens the step and may need more workers per rank. Declarations have finite domains
and a table budget; a new relation kind needs a server-side extension, and writing correct
relations is the application's task.

\section{Related Work}
\label{sec:related}

\textbf{Speculative decoding.} Speculative sampling~\cite{leviathan2023fast,chen2023accelerating}
and tree verification~\cite{miao2024specinfer,sun2023spectr,chen2024sequoia} draft from
learned heads~\cite{cai2024medusa,li2025eagle}, Jacobi iteration~\cite{fu2024break},
retrieval~\cite{he2024rest,oliaro2025suffixdecoding} or the prompt, and their gain falls
with the batch~\cite{liu2026speculative,zhang2025batch,wang2026agentspec}. Jump-forward
decoding~\cite{zheng2024sglang}, llguidance's forced tokens and
DOMINO~\cite{beurerkellner2024guiding} skip or speculate tokens a grammar determines, and
adaptive speculation tunes one draft length per step~\cite{hou2025banditspec}. \sys drafts
what relations across fields determine and sizes each request's verify by it, so
\cref{eq:width} divides a step's slots among requests knowing part of each return.

\textbf{Constrained decoding.} Grammar engines~\cite{dong2025xgrammar,li2026xgrammar,willard2023efficient}
and set automata~\cite{su2026vectorizing,xu2026trie} overlap bitmask computation with the
GPU. Semantic decoders check prefixes against completion
engines~\cite{scholak2021picard,poesia2022synchromesh}, static
analysis~\cite{agrawal2023monitor,mundler2025type}, composed
parsers~\cite{li2025correctness}, a live environment~\cite{zhang2026decode}, eager
constraints~\cite{beurerkellner2023prompting} or programmable
realizability~\cite{nagy2026chopchop}; their checks are local. Solver pipelines answer or
check outside the decoder~\cite{liu2023llmp,pan2023logiclm,kambhampati2024position}. \sys
enforces exact joint relations~\cite{chavan2026constrained} inside the decoder by moving the
heavier check to worker processes, builds on consistency and
elimination~\cite{freuder1982sufficient,beeri1983desirability,dechter1987network,dechter1999bucket,regin1994filtering}
as FactorDLM does for diffusion models~\cite{su2026constraints}, and composes with work on
masking's distortion~\cite{park2024grammar,nie2026future,reddy2026draft}.

\section{Conclusion}
\label{sec:conclusion}

Serving engines' drafters miss the text a structured record's relations determine, and
grammar masks enforce only syntax, so the model must satisfy the relations, which it
violates. \sys serves exact relational constraints as speculative drafts: an exact
feasibility check decides which tokens the relations determine, the constraint drafts them
as certain and sizes each request's verify by them, and worker processes compute the masks
while the GPU runs. Its drafts yield up to 10.5 tokens per verify step, and it matches or
outruns speculators tuned per load by up to 1.43$\times$, with every record valid at a
grammar mask's cost; the gain is largest where the relations compute the record. A
constraint that knows which tokens it determines is itself a drafter, so serving engines can
treat exactness as a source of speed.

\bibliography{references}
\bibliographystyle{mlsys2025}

\section*{AI Usage Disclosure}

We used Claude (Anthropic) and ChatGPT (OpenAI), large language model assistants, to edit the
writing of this paper: to proofread the text, correct grammar and spelling, and improve the wording, clarity
and consistency of its sentences. The authors reviewed every edit and take full
responsibility for the content of the paper.

\appendix

\section{Design details}
\label{app:design}

\Cref{alg:worker} gives the step of one constraint worker (\cref{sec:service}). A worker
answers every request open on it in one reply per step. Where the accepted tokens are the
determined run it computed ahead, the reply is already written; otherwise the worker
advances the session from as far as the computed-ahead state matches. After replying, it
computes ahead for every request whose sent tree is wholly determined, a chain of the full
node budget under a speculator or a single-token root otherwise, and whose next position
after it is determined too; it stops between two
field values when the next step's message arrives, so computing ahead always yields to the
next step.

\begin{algorithm}[h]
\caption{A worker's step.}
\label{alg:worker}
\begin{algorithmic}[1]
\REQUIRE one message per request open on the worker: its accepted tokens and its ask
\FOR{each request $r$ of the step}
  \IF{$r$'s accepted tokens are the run computed ahead}
    \STATE reply with the reply computed ahead
  \ELSE
    \STATE advance $r$'s session, from the computed-ahead state where it matches
    \STATE build the ask: the constraint's tree under the node budget, the rows of another
      drafter's tree, or the root row
    \STATE write each node's mask row into $r$'s free shared-memory block
  \ENDIF
\ENDFOR
\STATE send one reply for the step
\FOR{each request $r$ whose sent tree and next position are determined}
  \STATE build $r$'s next reply into its other block, one field value at a time
  \STATE stop between two values if the next step's message has arrived
\ENDFOR
\STATE poll for the next message for up to 1\,s, then block
\end{algorithmic}
\end{algorithm}

\Cref{alg:width} gives the per-request width allocation of \cref{sec:width}. Each request
starts at its root. Its certain nodes return one token each and come first; its guessed
nodes follow with the slopes of $g$'s least concave majorant, which are non-increasing, so
one pass
over all marginal returns in decreasing order visits the best allocation of every total
number of slots, and the allocation kept is the one with the most tokens per unit of step
time. A request whose tree has more guessed nodes than any verify has yet measured takes its
whole tree, and that verify measures every cut of it.

The engine's verify bookkeeping keeps its uniform layout at the widest request's width,
with a request's slots past its width unreachable, while the forward pass computes the real
slots alone: each request gets a separate query length and packed tree mask. In this mode
the verify graphs, captured per batch size at widths 2, 4, 8 and the server's, run a step
padded to the narrowest that holds it. Under tensor parallelism each rank's service builds
trees on a separate schedule, so the ranks compare a hash of each step's trees and follow the
first rank's where they differ.

\begin{algorithm}[h]
\caption{Per-request verify widths (\cref{eq:width}).}
\label{alg:width}
\begin{algorithmic}[1]
\REQUIRE each request's tree, its first $c_r$ nodes certain; slopes
  $\hat g_1 \ge \hat g_2 \ge \dots$ of $g$'s least concave majorant; step cost
  $S(T) = a + b \max(K, F(T))$
\STATE $W_r \gets 1$ for each request $r$; $R \gets T \gets$ the number of requests
\STATE $M \gets$ for each $r$: $c_r$ returns of 1, then $\hat g_1, \hat g_2, \dots$ for its
  guessed nodes
\STATE $W^\star \gets W$; $\rho^\star \gets R / S(T)$
\FOR{each $(r, m) \in M$ by decreasing $m$}
  \STATE $W_r \gets W_r + 1$; $R \gets R + m$; $T \gets T + 1$
  \IF{$R / S(T) > \rho^\star$}
    \STATE $W^\star \gets W$; $\rho^\star \gets R / S(T)$
  \ENDIF
\ENDFOR
\STATE \textbf{return} $W^\star$; request $r$ verifies its tree's first $W^\star_r$ nodes
\end{algorithmic}
\end{algorithm}

\section{Evaluation details}
\label{app:setup}

\textbf{Platform.} Two NVIDIA RTX A6000 GPUs with 48\,GB each, and a two-socket Intel Xeon
Gold 6338 host with 64 cores and 1\,TB of memory. Qwen3-8B~\cite{qwen2025qwen3} with the
AngelSlim EAGLE-3 head, Llama 3.1 8B Instruct~\cite{llama2024llama} with SGLang's
packaging of the EAGLE-3 authors' head, Qwen3-1.7B and Qwen3-14B run in bfloat16 with a
12,288-token context and Qwen3's thinking mode off unless stated. SGLang 0.5.16 and vLLM
0.21.0 are each capped at 30\,GiB of device memory per GPU.

\textbf{Workloads.} The \emph{ledger} is a synthetic tool-calling record whose transfers
must refer to accounts the record declares, with distinct endpoints and a reversal that
copies an earlier transfer. $\tau$-bench retail and BFCL multi-turn calls must name the
orders, items and files the conversation declares; the retail relations were mined from the
training split, and we evaluate on the test split. The planning workloads come from NATURAL
PLAN, excluding the problem each prompt uses as its worked example: 999 trip problems of three
to seven cities, all 1,599 of three to ten as a full set, and 999 meeting and 999 calendar
problems. An itinerary must visit every city for its stated days over direct flights within
the event windows. A meeting schedule must respect travel times and availability windows
and meet the most friends any plan meets; the model chooses whom and when. A calendar slot
must fit every busy interval; on 600 problems several slots do, and the model chooses by the
stated preferences. Every prompt shows one worked example and asks for a JSON record, and
the meeting prompt states how many friends the best plan meets.

\textbf{Tuned shapes.} \Cref{tab:shapes} lists the fastest shape of each speculator per
cell at 8 and 64 in flight; the tuned speculator of every cell is the n-gram drafter's. At
128 in flight the speculator is the best of ten shapes, since width 32's verify exceeds the
attention kernel's workspace at that load.

\begin{table}[h]
\centering
\caption{The fastest shape of each speculator per cell, Qwen3-8B on one A6000: the
n-gram drafter's width, and EAGLE-3's tree as steps by ways or a chain.}
\label{tab:shapes}
\footnotesize
\setlength{\tabcolsep}{4pt}
\begin{tabular}{@{}lllll@{}}
\toprule
& \multicolumn{2}{c}{8 in flight} & \multicolumn{2}{c}{64 in flight} \\
\cmidrule(lr){2-3}\cmidrule(l){4-5}
Workload & n-gram & EAGLE-3 & n-gram & EAGLE-3 \\
\midrule
Ledger & 16 & $5{\times}4$ & 8 & chain 4 \\
$\tau$-retail & 12 & chain 4 & 8 & chain 4 \\
BFCL & 12 & default & 4 & chain 4 \\
Trip & 16 & $5{\times}4$ & 8 & chain 4 \\
Meeting & 16 & $5{\times}4$ & 8 & chain 4 \\
Calendar & 16 & $5{\times}4$ & 4 & chain 4 \\
\bottomrule
\end{tabular}
\end{table}

\textbf{Baselines per setting.} In vLLM the baselines are \xg, the n-gram drafter at draft
lengths 2, 4, 8 and 16, and EAGLE-3 at three draft tokens. In SGLang, the settings over two
GPUs and on Qwen3-1.7B and Qwen3-14B compare against \xg and the n-gram drafter at widths 4,
8, 12 and 16; on Llama 3.1 8B also against EAGLE-3 at its default tree and as chains of two
and four; and at temperature one against \xg and the n-gram drafter at width 12. The
settings over two GPUs, on Qwen3-1.7B and Qwen3-14B and at temperature one ran at a later
commit than their baselines; their lead passes through \xg, measured in both sessions, so a
change in the host's speed cancels.

\textbf{Requests per cell.} Closed loop serves 64 requests per cell up to 16 in flight and
192 above, 192 also at 8 in flight in the tuned, solver and notes runs, and 256 at 128.

\textbf{Enumerated schemas.} Every calendar and retail record fits under the cap of 1,024
feasible assignments, in at most 16 and 31 \code{anyOf} branches once merged; 381 of the 384
BFCL calls fit, in at most 49 branches, and the client refuses the other 3. Compiling a
schema on the client takes at most 0.9\,s.

\textbf{Thinking.} Qwen3 reasons unmasked until \code{</think>} or a budget of 8,192
tokens, which with the longest prompt, 2,140 tokens on meeting planning, and the record stays
inside the 12,288-token context; \xg then constrains the record. A trace that reaches the
budget is closed by the engine and its record still decoded and scored. With this budget
the $\tau$-bench prompts exceed the context, so thinking serves the planning workloads.

\section{Additional results}
\label{app:results}

\begin{figure}[h]
\centering
\includegraphics{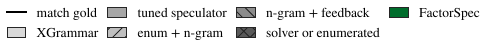}\\
\includegraphics{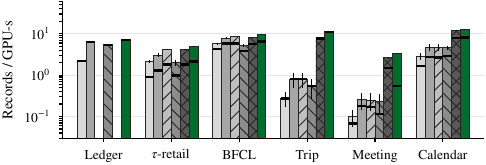}
\caption{Valid records per GPU-second (bars, log scale) and records that also match gold
(marks) at 8 in flight, with the baselines of \cref{fig:valid}; 95\% intervals over prompts
and runs.}
\label{fig:valid8}
\end{figure}

\textbf{Valid records at 8 in flight.} \Cref{fig:valid8} repeats \cref{fig:valid} at 8 in
flight, where the leads of RQ3 hold: 13.9$\times$ on trip, 13.2$\times$ on meeting and
2.6$\times$ on calendar planning over the syntax-only baselines.

\textbf{What the exact check costs.} \Cref{tab:querycost} times the work a worker does, on
one core, over 200 gold records per workload: each position's mask, the allowed set and
the advance by the gold token, and each feasibility query. Between records the heap is
collected and frozen as a worker does when a request ends, so every timed query excludes a
full garbage collection. The memo column answers each distinct query of the first 50
records again on a fresh engine, whose memos start empty, and gives the ratio of the two
totals.

\begin{table}[h]
\centering
\caption{The exact check's cost per position and per query: medians in $\mu$s, 99th
percentiles and the slowest query in ms, and how much cheaper the memos make the distinct
queries. Below, trip planning over three to ten cities, by city count.}
\label{tab:querycost}
\footnotesize
\setlength{\tabcolsep}{3.5pt}
\begin{tabular}{@{}lrrrrrr@{}}
\toprule
& \multicolumn{2}{c}{Position} & \multicolumn{3}{c}{Query} & \\
\cmidrule(lr){2-3}\cmidrule(lr){4-6}
Workload & p50 & p99 & p50 & p99 & max & Memo \\
\midrule
Ledger & 4.7 & 0.79 & 2.1 & 0.19 & 0.53 & 1.6$\times$ \\
$\tau$-retail & 4.3 & 1.95 & 75.7 & 0.76 & 1.07 & 2.5$\times$ \\
BFCL & 4.9 & 2.17 & 16.1 & 0.13 & 0.20 & 1.5$\times$ \\
Trip & 4.5 & 0.48 & 130.5 & 0.55 & 1.03 & 2.1$\times$ \\
Meeting & 4.3 & 10.22 & 25.0 & 0.06 & 3.22 & 1.1$\times$ \\
Calendar & 4.5 & 0.38 & 54.9 & 0.08 & 0.11 & 0.9$\times$ \\
\midrule
3 cities & 4.6 & 0.35 & 100.6 & 0.22 & 0.23 & 1.6$\times$ \\
4 cities & 4.5 & 0.39 & 115.0 & 0.29 & 0.31 & 1.9$\times$ \\
5 cities & 4.5 & 0.41 & 130.2 & 0.36 & 0.41 & 2.0$\times$ \\
6 cities & 4.5 & 0.47 & 138.4 & 0.44 & 0.48 & 2.2$\times$ \\
7 cities & 4.5 & 0.52 & 153.6 & 0.57 & 0.60 & 2.3$\times$ \\
8 cities & 4.5 & 0.67 & 163.8 & 0.69 & 1.37 & 2.5$\times$ \\
9 cities & 4.5 & 0.84 & 179.5 & 0.95 & 1.78 & 2.4$\times$ \\
10 cities & 4.6 & 1.31 & 198.4 & 1.71 & 2.61 & 2.7$\times$ \\
\bottomrule
\end{tabular}
\end{table}

\begin{figure}[h]
\centering
\includegraphics{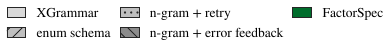}\\
\includegraphics{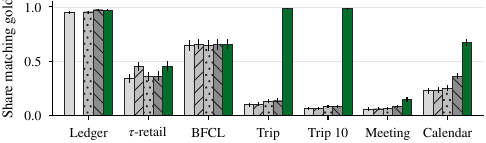}
\caption{Over every full set \sys matches gold far more often on planning, and as often as
the best baseline on tool calls. At 64 in flight, \sys at one width per step; Trip 10 is
all 1,599 trip problems; enum schemas on every workload except the ledger; 95\% Wilson intervals.}
\label{fig:full}
\end{figure}

\textbf{Matching gold over the full sets.} \Cref{fig:full} gives the share of each full set
whose record matches gold, beside the validity of \cref{fig:motivation}.

\begin{figure*}[h]
\centering
\includegraphics{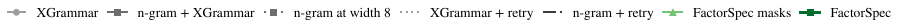}\\
\subfloat[Trip, by time]{\includegraphics{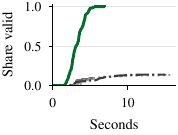}}%
\subfloat[Meeting, by time]{\includegraphics{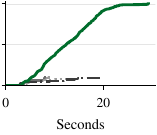}}%
\subfloat[Ledger, in 10\,s]{\includegraphics{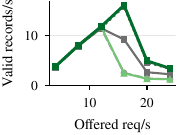}}%
\subfloat[Trip, in 10\,s]{\includegraphics{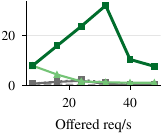}}%
\subfloat[Meeting, in 20\,s]{\includegraphics{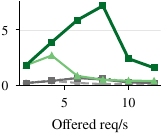}}%
\subfloat[Trip, p99]{\includegraphics{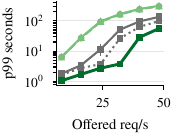}}
\caption{\sys answers every request validly and, past the grammar engine's capacity,
completes the most valid planning records inside the limit at the lowest tail latency.
(a, b) Share of requests valid by each time at 64 in flight, retries included. (c--f)
Poisson arrivals for two minutes per rate.}
\label{fig:time}
\end{figure*}

\textbf{Time to a valid record and open loop.} \Cref{fig:time} gives the time to a valid
record on trip and meeting planning and the service under Poisson arrivals on the ledger,
trip and meeting planning, which RQ3 quotes.

\begin{figure*}[h]
\centering
\includegraphics{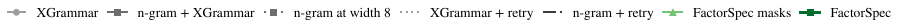}\\
\subfloat[Ledger, by time]{\includegraphics{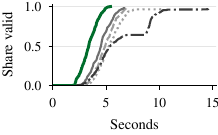}}%
\subfloat[$\tau$-retail, by time]{\includegraphics{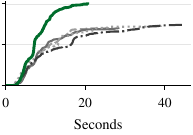}}%
\subfloat[BFCL, by time]{\includegraphics{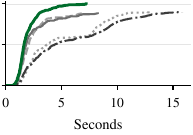}}%
\subfloat[Ledger, p99]{\includegraphics{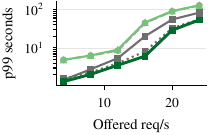}}%
\subfloat[Meeting, p99]{\includegraphics{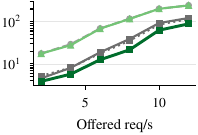}}
\caption{(a--c) The share of requests valid by each time at 64 in flight, retries included.
(d, e) 99th-percentile latency under Poisson arrivals, two minutes per rate.}
\label{fig:time-more}
\end{figure*}

On the tool calls, where most records are
valid under every configuration, the time to a valid record follows each configuration's
speed (\cref{fig:time-more}(a--c)). Under the open loop, \sys's 99th-percentile latency is
the lowest at every rate on the ledger and on meeting planning (d, e).

\textbf{A wider verify.} The server's widest verify is 12 slots. \Cref{tab:wide} compares
jump-forward, which verifies only the determined run, at 12 and 32 tokens per step, and the
factor drafter at up to 32 slots, each over the factor drafter at up to 12, in a separate
campaign whose servers shared one history. On trip planning at 8 in flight, a record the
relations fix whole, 32 slots run 1.20$\times$ faster; elsewhere they change the factor
drafter's throughput by $-$11 to $+$2\%. Jump-forward runs at 0.94 to 1.01 of the factor drafter when
both use 12 slots, and at 0.87 to 1.03 of it when both use 32; in this campaign the guessed nodes return less than the
4 to 12\% of RQ4.

\begin{table}[h]
\centering
\caption{Throughput over the factor drafter at up to 12 slots, by the verify's slots;
mean of three runs.}
\label{tab:wide}
\footnotesize
\setlength{\tabcolsep}{4pt}
\begin{tabular}{@{}lrrrr@{}}
\toprule
& & \multicolumn{2}{c}{Jump-forward} & Factor \\
\cmidrule(lr){3-4}\cmidrule(l){5-5}
Workload & In flight & 12 & 32 & 32 \\
\midrule
Ledger & 8 & 0.95 & 0.92 & 0.89 \\
Ledger & 64 & 0.97 & 0.96 & 0.98 \\
Trip & 8 & 1.00 & 1.19 & 1.20 \\
Trip & 64 & 1.01 & 1.03 & 1.02 \\
Meeting & 8 & 0.94 & 0.87 & 1.00 \\
Meeting & 64 & 0.94 & 0.91 & 0.95 \\
\bottomrule
\end{tabular}
\end{table}

\end{document}